\documentclass[conference]{IEEEtran}
\usepackage{latexsym}
\usepackage{adjustbox}
\usepackage{multicol, multirow}
\usepackage{hyperref}
\usepackage{diagbox}
\usepackage{afterpage}
\usepackage{cite}
\usepackage{amsmath,amssymb,amsfonts, bbm, array}
\usepackage{algorithmic}
\usepackage{graphicx}
\usepackage{textcomp}
\usepackage{xcolor}
\usepackage[utf8]{inputenc}
\usepackage[T1]{fontenc}
\def\BibTeX{{\rm B\kern-.05em{\sc i\kern-.025em b}\kern-.08em
    T\kern-.1667em\lower.7ex\hbox{E}\kern-.125emX}}
\begin{document}

\title{Towards Model-Based Data Acquisition for Subjective Multi-Task NLP Problems}


\author{\IEEEauthorblockN{
Kamil Kanclerz,
Julita Bielaniewicz,
Marcin Gruza, \\
Jan Kocoń, 
Stanisław Woźniak, and
Przemysław Kazienko}
\IEEEauthorblockA{\textit{Department of Artificial Intelligence, Wrocław University of Science and Technology, Poland}}
\IEEEauthorblockA{\footnotesize{\texttt{\{kamil.kanclerz, julita.bielaniewicz, marcin.gruza,jan.kocon,}}} 
\IEEEauthorblockA{\footnotesize{\texttt{stanislaw.wozniak, kazienko\}@pwr.edu.pl}}}
}


\maketitle

\begin{abstract}
Data annotated by humans is a source of knowledge by describing the peculiarities of the problem and therefore fueling the decision process of the trained model.
Unfortunately, the annotation process for subjective natural language processing (NLP) problems like offensiveness or emotion detection is often very expensive and time-consuming. 
One of the inevitable risks is to spend some of the funds and annotator effort on annotations that do not provide any additional knowledge about the specific task. 
To minimize these costs, we propose a new model-based approach that allows the selection of tasks annotated individually for each text in a multi-task scenario. 
The experiments carried out on three datasets, dozens of NLP tasks, and thousands of annotations show that our method allows up to 40\% reduction in the number of annotations with negligible loss of knowledge. The results also emphasize the need to collect a diverse amount of data required to efficiently train a model, depending on the subjectivity of the annotation task. We also focused on measuring the relation between subjective tasks by evaluating the model in single-task and multi-task scenarios. Moreover, for some datasets, training only on the labels predicted by our model improved the efficiency of task selection as a self-supervised learning regularization technique.
\end{abstract}

\begin{IEEEkeywords}
natural language processing, personalization, self-supervised learning, data acquisition, model-based annotation optimization
\end{IEEEkeywords}

\section{Introduction}
\label{sec:introduction}
One of the most crucial parts of developing any machine learning solution is the data acquisition process. A well-prepared dataset will significantly increase the amount of knowledge obtained by the model during training. However, in most cases, obtaining a high-quality dataset includes a large annotation process and further data post-processing, including filtering out a considerable part of the dataset. In this way, the financial expenses and time required to collect the filtered data are wasted. The most common approaches to tackle this problem focus on maximizing the inter-annotator agreement or selection of texts, which should further improve the model performance. However, these methods assumed the existence of only one true label for a text. Additionally, data annotation optimization techniques allowed for including or fully omitting the specific text during the annotation process. This led to the loss of valuable knowledge that could be extracted from the subset of labels for the omitted text. In addition, some labels of texts selected for the annotation procedure may not provide additional knowledge to the model and be just a waste of time and money.

To the best of our knowledge, currently, there are no methods designed for subjective multi-task NLP tasks, which focus on the above issues. Therefore, we present our novel model-based data acquisition strategy, which operates on the level of individual labels and allows the user to annotate only a subset of labels for a specific text while providing automatic annotations for the rest of the labels. To evaluate our method, we performed a complex evaluation that included several experimental scenarios. We also developed our own measures and used them alongside commonly used ones to better verify the effectiveness and reliability of the proposed technique. 

The main contributions of this work are as follows: (1) we proposed a novel model-based data acquisition optimization strategy focused on reducing the annotation effort by predicting the valuable labels for each text resulting in up to 25\% benefit and up to 40\% reduction of the annotation effort (Fig.~\ref{fig:model_based_data_acquisition}); (2) we applied our method on three datasets regarding personalized multi-task NLP problems; (3) we developed new evaluation metrics appropriate for the problem and leveraged them along with the standard ones like macro F1-score; (4) our evaluation included the self-supervised scenario, where the model was trained only on labels previously predicted by itself to measure the amount of knowledge not learned during the training procedure; (5) we also tested the impact of training dataset size on the model performance; (6) we analyzed the knowledge transfer between tasks in single-task and multi-task scenario; (7) we also analyzed the relation between the number of annotations per single text, number of unique texts in the dataset, and the model performance; (8) we conducted the evaluation of the personalized architecture for multi-task subjective problems.

\section{Related Work}
\label{sec:related_work}
Ever since the very beginning of research related to artificial intelligence, there has been a consistent series of issues regarding data acquisition. A crucial part of the said possible hardships include estimating the number of information needed for a reasonable analysis and, of course, the costs needed for obtaining it. There is no doubt that without adequate quality and quantity of data, we will certainly omit important information regarding relations between data. Moreover, such miscalculation may lead to false conclusions that will distort the message of a number of studies and, consequently, the entire area of study. Such oversight must be prevented by all means, and thus researchers must find the balance between quality and expense.

\subsection{Data Acquisition}
There are a considerable number of studies that specialize in the area of data acquisition, most of which revolve around the generalization approach to data analysis. The work \cite{brain1999effect} foresees an accurate image of the current state of data acquisition. Managing large amounts of data comes down to counting losses we are ready to sacrifice, and finding balance is an approach where the authors reduced the size of a training test with the cost of decreasing variance.
Cantrell \cite{cantrell2007methodological} suggests that online data collection methods provide an advantage in the possibility of using online tools, but in his works it is not reconsidered whether the costs of the quantity approach may have been avoided to some extent.

\subsection{Dataset Distillation}
An interesting measure of quality was introduced in the work \cite{hutchinson2010resident} that signals the level of quality of a chosen dataset. This creates an opportunity to calculate the metric each time we reduce the amount of data; however, when faced with ad hoc analysis, there is a strong bias to the quantity that overshadows the data quality.
The article \cite{rose2011garbage} presents an empirical view on the matter, implying that certain fields in which data are collected are burdened with the natural impossibility of acquiring quality datasets. Thus, since data quality tools are as good as the data collected, it prevents certain areas of study from having a proper analysis. This approach requires the use of a set of metrics that can guarantee high-quality, considerable quantity data, regardless of the field of study.
In the book \cite{cavanillas2016new} there is a carefully conducted analysis of possible approaches to dataset management and methods for extracting quality data from available sources. An article \cite{lyko2016big} is especially interesting because it provides insight into the data acquisition process used in large corporations. Although efficient, they seem to be too strict in avoiding quantity bias, discarding a lot of useful data in the process.
An approach presented in the article \cite{wang2018dataset} aims to receive a distilled version of datasets through the use of descending gradient and different initialization techniques. Although promising, it only works on very simple datasets, struggling when faced with multidimensional scenarios.
D. Barrett \cite{barrett2018data} argues that improving data collection techniques is the key to having a distilled data set from the beginning. This study aims to focus on certain methods for obtaining quality data, which is clearly an engaging process, but we disagree with omitting acquisition beyond the data collection process.
In the work \cite{nguyen2021dataset} the authors perform a dataset distillation using tailored algorithms applied to
convolutional architectures, which results in interesting enhancements of the distilled data, but additional analyses and explorations would be needed to provide insight into the full potential of the presented methods.
The article \cite{vidgen2020directions} examines the task of data acquisition by conducting a systematic review of publicly available datasets for the detection of abusive content, focusing on improvements in training datasets when distilling data. 
The authors of article \cite{sucholutsky2021soft} focus on the multimodal nature of the distillation process and try to find the right balance for techniques that specialize in multidomain problems. This approach performs well in the multi-task approach, but it comes with the cost of lower performance values per modality.
The data acquisition researched in the article \cite{zhoudataset} aimed to receive a distilled dataset with reduced memory size and an improved training time using feature regression. The intriguing research proves to generalize well on different types of image datasets, but would unfortunately not work on the textual data.
The work \cite{liudataset} introduces a data distillation technique that utilizes factorization in order to separate the dataset into two, analyzing groups of hallucination values and the base values. One of the key advantages of this approach is the small number of hyperparameters needed for good results, but the method does not perform well if the time factor is crucial.
The article \cite{larasati2022review} provides a summary of the dataset distillation-based solutions to deep learning tasks with a focus on quality measurement after distillation. The authors do come to the conclusion that massive image datasets for image classification are vastly optimized for acquisition purposes, but the same assumption cannot be applied to the textual datasets.
When optimizing the initial and target network parameters for large-scale datasets, the authors of the work \cite{cazenavette2022dataset} compute and store training trajectories of expert networks. Although it outperforms many available methods, it comes with the cost of additional computational costs that may not be possible to achieve for many researchers.
The authors of the work \cite{sachdeva2023data} present a formal framework for data distillation, along with providing a detailed taxonomy of existing approaches with respect to multi-task data. Although many advanced methods for annotation process optimization have been developed in the field of computer vision, no similar advances have been made in the field of natural language processing, which implies a promising future for the area of textual data acquisition.
The article \cite{zhang2023balanced} proposes a student-teacher network that participated in the data acquisition process in long-tailed scenarios. The introduced framework benefits from sample diversity and learns generalized representation, which may indicate the possible area of personalization tasks to be enhanced in the future.
The authors of article \cite{radenovic2023filtering} propose a straightforward filtering strategy that significantly reduces the size of the dataset and achieves improved performance across zero-shot vision-language tasks. The prominent disadvantage can be noticed in noisy datasets, as there is a certain loss of performance when faced with preprocessed datasets.

\subsection{Label Distillation}
Another promising work \cite{bohdal2020flexible} implies that it is the labels that should be distilled instead of the data. The authors of the article focus on the crafting of synthetic labels for arbitrarily chosen standard data, which works in analyzed research, but can fail to perform in a different area of study than the examined one. 
The work \cite{gao2023forget} investigates a new crowd counting task in an incremental domain training setting using a single model updated by the incremental domains. The method is interesting and performs well when not dealing with missing annotations, as this scenario heavily burdens the model, resulting in a possible decline in performance.
The authors of the survey \cite{whang2023data} study the research scope for data collection and data quality primarily
for deep learning applications with a special focus on bias and fairness of data distillation. It is especially emphasized that noisy or missing labels cause poor generalization of the test data and this implies the potential for personalization research.

\subsection{Model-based Techniques}
The authors of the article \cite{chen2023sssd} propose model-based self-supervised self-distillation methods that extract representations of the target dataset and generate pseudo labels through clustering. The downside of this technique is mostly the lack of adaptability for textual data, otherwise very promising in future directions of data acquisition.

\subsection{Research Gap}
When it comes to textual data, there is a prominent lack of research when facing the data acquisition process, especially in the domain of personalization \cite{KANCLERZ2020128,milkowski2021personal,kocon2021learning,kocon2021ipm,10.1007/978-3-030-77964-1_24,ngo2022studemo,milkowski2022multitask,kanclerz2022if,bielaniewicz2022deep,ferdinan2023personalized,mieleszczenko2023capturing,kocon2023differential,kazienko2023human,kocon2023chatgpt}. In this work, 
we have focused on the matter of developing a set of measures that help evaluate the overall quality of textual data after the distillation process, which also performs well in the personalization scenario.

\begin{figure*}
\centering
\includegraphics[width=1.0\linewidth]{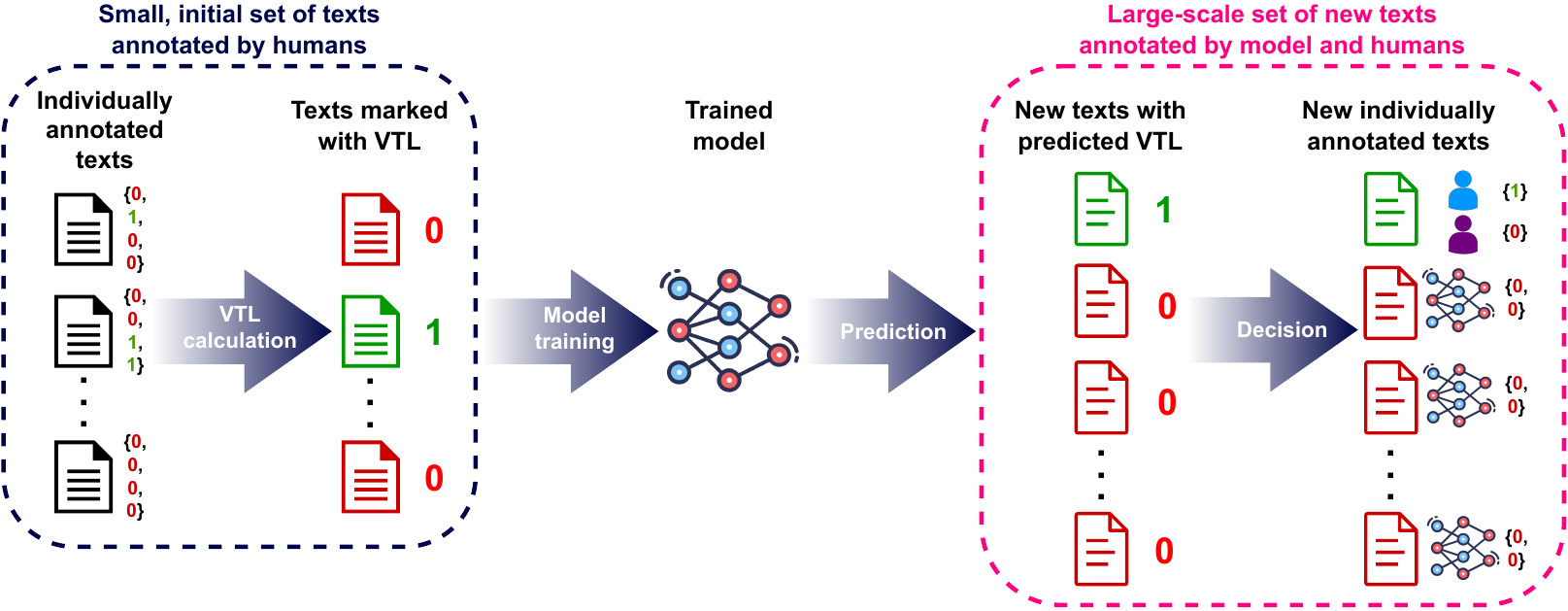}
\caption{Our model-based labeled data acquisition schema with the use of the Valuable Text Label (VTL) metric, presented for a single label (one subjective NLP problem/task).} 
\label{fig:model_based_data_acquisition}
\end{figure*}

\begin{figure}
\centering
\includegraphics[width=1.0\linewidth]{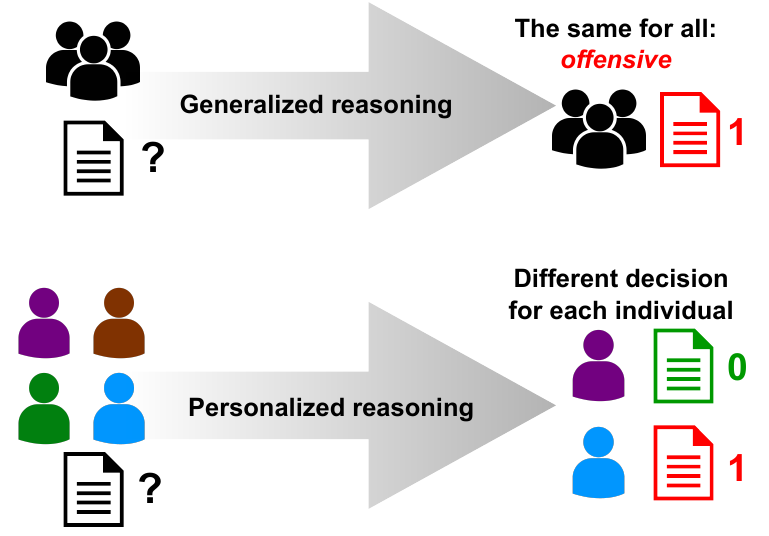}
\caption{Difference between the generalized and personalized approach in offensiveness detection.} 
\label{fig:generalized_vs_personalized}
\end{figure}
\section{Model-Based Data Acquisition for Subjective NLP Problems}
\label{sec:model_based_data_acquisition_for_subjective_nlp_problems}

\subsection{Subjective NLP Tasks}
\label{sec:subjective_nlp_problems}
The variety of problems in the field of natural language processing is generously wide, so much so that in the perspective of just the group of subjective problems, there is an area of many possibilities for an individual understanding of the many of said perspectives are still yet to be discovered, as each person can interpret a single information very differently. The usual approach in the dominant number of studies across all NLP fields focuses mainly on the majority of annotators. This not only discriminates against people who do not tend to follow trends, but also excludes a significant amount of useful data that could otherwise broaden the perspective of certain peculiarities and tendencies, as presented in Fig.~\ref{fig:generalized_vs_personalized}.
In our case, we simultaneously consider multiple subjective problems (labels) such as emotions, offensiveness, irony, and humor for each text, up to 23 labels (Sec.~\ref{sec:doccano_dataset}). This approach not only implies a wider range of analysis per user, but also hints at a much broader grasp of relations between each dimension.

\subsection{Problem Description}
\label{sec:problem_description}

In almost all subjective tasks in NLP, there are many texts being collected and found by all or almost all people irrelevant to the problem, e.g. \textit{not funny} in humor detection. This problem is even more crucial if we want to annotate texts simultaneously for many not related problems (multi-task), e.g. \textit{sadness} and \textit{funniness}. If a given text $d$ is annotated by many humans as \textit{sad}, it is unlikely to be annotated by a significant fraction of the same group of people as \textit{funny}. It means that asking for annotation of $d$ with \textit{funny} would provide only or almost only neutral labels (zeros). Then, we should avoid such useless annotations. 
To identify such cases, we have developed a model-based procedure for data acquisition, Sec.~\ref{sec:model_based_data_acquisition}.
Having the labeled data acquisition model, we can estimate its quality using appropriate measures (Sec.~\ref{sec:annotation_effort_reduction} - \ref{sec:model_benefit}). We also exploited the received labeled data in the real models solving subjective problems in a personalized and multi-task setup, i.e. prediction of all labels $l \in L$ for a given text $d$ and individual reader (Sec.~\ref{sec:self_supervised_model_evaluation}).

\subsection{Valuable Text Label (VTL)}
\label{sec:valuable_text_label}

To estimate how valuable the label $l$ (problem, task) should be considered for text $d$, we developed a new measure called the Valuable Text Label (VTL). It divides the number of non-zero annotations ($a \not\in \{0\}$) by the number of all annotations $|A_{d, l}|$ with respect to the specified label $l$ in the context of a certain text $d$. The calculated value is then compared with the threshold $t$, which takes values in the range $[0, 1]$. The $VTL(d, l, t)$ measure returns 1, if the computed value is greater than or equal to the threshold $t$, and 0 otherwise. VTL quantifies whether the text $d$ has the potential to be suitable for at least $t$ percentage of the population to find it relevant to the task $l$. For example, text $d$ that no one or very few people find \textit{funny} should not be annotated with the $l=funny$ label. We assume that collecting labels for text $d$ with only or almost only $l=0$ does not provide any useful knowledge. The measure value is calculated as follows:

\begin{equation}
VTL(d, l, t) = \begin{cases}
                            1, & \text{if } \frac{\sum_{a \in A_{d, l}} \mathbbm{1}_{\{a \not\in \{0\}\}}}{|A_{d, l}|} \geq t \\
                            0, & \text{otherwise}
                      \end{cases}
\label{eq:valuable_text_label}
\end{equation}

\subsection{Model-Based Data Acquisition}
\label{sec:model_based_data_acquisition}

To leverage the possibilities of deep neural architectures in the data acquisition process, we developed model-based data acquisition that is used to annotate texts and identify those of them that are valuable for further human annotations, Fig.~\ref{fig:model_based_data_acquisition}.

We start with the preprocessing of the relatively small initial set of multiple annotations previously collected from humans. It is used to train our model and quantify how valuable each label $l$ is in the context of the specific text $d$. For that case, we use the $VTL(d,l,t)$ measure described in Sec.~\ref{sec:valuable_text_label}. The output $\hat{y}$ of our model is the predicted value of $=VTL(d,l,t)$ for a given text $d$ and task $l$ that is directly used for model-based annotations or pre-selection for human annotations.

After label preprocessing, we train the deep neural network on the texts annotated with the VTL value for each label. The obtained model is used to predict the values of our metric for each new text (candidates). The VTL values received from the model are used to decide whether the label should be annotated by 
humans
($\hat{y} \in \{1\}$) or should it be done automatically by the model ($\hat{y} \in \{0\}$). If the model recommends human annotations, i.e., its predicted value of $VTL(d,l,t)=1$, then a given text $d$ is labeled by all humans for the task $l$ to capture individual peculiarities in human text perception (subjectivity of the task), i.e., expected controversy of the text and conformity of users \cite{kanclerz-etal-2021-controversy}.

In other words, if the model predicts $VTL(d,l_1,t)=0$ and $VTL(d,l_2,t)=1$, then the text $d$ is annotated by the model with $l_1=0$ for all users, and the annotators manually annotate $d$ with the task $l_2$. The lack of human annotation of $d$ with $l_1$ is our gain, as $l_1=0$ has been achieved without any human involvement. In this way, we are able to select texts that are more relevant to individual tasks, which is very important in multi-task scenarios\footnote{See Sec.~\ref{sec:doccano_dataset} with our Doccano dataset that contains 23 simultaneously acquired tasks.}.

\subsection{Annotation Effort Reduction (AER)}
\label{sec:annotation_effort_reduction}

Our novel model-based method assumes that only valuable text labels are worth being annotated by annotators. This means that labels $l$, which are invaluable for a specific text $d$ according to the threshold $t$, should be automatically marked with the $\{0\}$ class and skipped during the annotation process with real users. To efficiently measure the effort reduction caused by our approach, we developed the Annotation Effort Reduction (AER) metric. It counts invaluable labels ($VTL(d, l, t) \in \{0\}$), which were correctly predicted by our model ($VTL(d,l,t) = \hat{y}_{d,l}$) across all labels $l \in L$ and all texts $d \in D$. The calculated value is then divided by the number of all possible annotations, which is equal to the number of texts in the dataset ($|D|$) multiplied by the number of all possible labels ($|L|$). The result of the division is the percentage amount of the reduced annotation effort. The exact formula used for computing $AER(D,\hat{y},t)$ is presented in Eq.~\ref{eq:annotation_effort_reduction}.

\begin{equation}
AER(D, \hat{y}, t) = \frac{\sum_{d \in D} \sum_{l \in L} \mathbbm{1}_{\{VTL(d,l,t) \in \{0\} \land VTL(d, l, t) = \hat{y}_{d, l}\}}}{|D| * |L|}
\label{eq:annotation_effort_reduction}
\end{equation}

\subsection{Absolute Annotation Loss (AAL)}
\label{sec:absolute_annotation_loss}

Relying on the entire process on whether a specific label $l$ is valuable for text $d$ only on the model predictions $\hat{y}$ carries the risk of skipping labels that may in fact turn out to be important. To measure the possible loss of useful labels, we developed the Absolute Annotation Loss (AAL) metric. It calculates the number of valuable labels ($VTL(d, l, t) \in \{1\}$), for which the predictions of our model were wrong ($VTL(d,l,t)\neq\hat{y}_{d,l}$). This operation is conducted across all labels $l \in L$ and on all texts $d \in D$. Then, the computed value is divided by the number of all important labels in the dataset ($VTL(d,l,t) \in \{1\}$). The formula for calculating $AAL(D, \hat{y}, t)$ is following:
\begin{equation}
AAL(D, \hat{y}, t) = \frac{\sum_{d \in D} \sum_{l \in L} \mathbbm{1}_{\{VTL(d,l,t) \in \{1\} \land VTL(d, l, t)\neq\hat{y}_{d, l}\}}}{\sum_{d \in D} \sum_{l \in L} \mathbbm{1}_{\{VTL(d,l,t) \in \{1\}\}}}
\label{eq:absolute_annotation_loss}
\end{equation}
\subsection{Mean Label Rarity Annotation Loss (MLRAL)}
\label{sec:mean_label_annotation_loss}

To measure the possible loss of information caused by our model with respect to the distribution of each label, we propose the Mean Label Rarity Annotation Loss (MLRAL) metric. In the first step, it calculates the Label Annotation Loss (LAL) 
for each of the possible labels $l$ separately. The $LAL(D,l,\hat{y},t)$ computes the percentage value of valuable labels ($VTL(d, l, t) \in \{1\}$), for which the model made the incorrect decision ($VTL(d,l,t)\neq\hat{y}_{d,l}$). The calculated value is further divided by the number of all samples $d \in D$, for which the specific label $l$ is considered useful ($VTL(d,l,t) \in \{1\}$):

\begin{equation}
LAL(D, l, \hat{y}, t) = \frac{\sum_{d \in D}  \mathbbm{1}_{\{VTL(d,l,t) \in \{1\} \land VTL(d, l, t)\neq\hat{y}_{d, l}\}}}{\sum_{d \in D} \mathbbm{1}_{\{VTL(d,l,t) \in \{1\}\}}}
\label{eq:label_annotation_loss}
\end{equation}

In the next step, the $LAL$ metric values are averaged across all possible labels $l \in L$:

\begin{equation}
MLRAL(D, \hat{y}, t) = \frac{\sum_{l \in L} LAL(D, l, \hat{y}, t)}{|L|}
\label{eq:mean_label_annotation_loss}
\end{equation}

\subsection{Model Benefit (MB)}
\label{sec:model_benefit}

To measure and better interpret the advantage of applying our model-based approach in the data acquisition process, we developed the Model Benefit (MB) metric. It is the difference between the gain defined by $AER(D,\hat{y},t)$ and the knowledge loss calculated by $AAL(D,\hat{y},t)$ for a specific dataset $D$, the predictions of the model $\hat{y}$, and the threshold $t$:

\begin{equation}
MB(D, \hat{y}, t) = AER(D, \hat{y}, t) - AAL(D, \hat{y}, t)
\label{eq:model_benefit}
\end{equation}

The positive value of $MB(D, \hat{y}, t)$ indicates that the reduction of the annotation effort was greater than the loss of knowledge caused by incorrect model predictions. On the contrary, the negative MB value means that the loss of knowledge affects the greater part of the dataset than the one acquired automatically through model predictions. 

\section{Datasets}
\label{sec:datasets}
The great importance of data used during our experiments was a key element in obtaining genuine results. We needed to accumulate data that were sufficiently diverse so that each subjective problem was adequately represented. For this reason, we have launched a project named Doccano 1.0, where individuals annotated a diverse number of texts that corresponded to subjective problems. Furthermore, after a thorough analysis of many datasets, we have also chosen to expand our experimental set of data by adding two sources, Measuring Hate Speech and Unhealthy Conversations. Although the volume and quality of the data were undoubtedly crucial, the vast difference between the datasets is prominent to a degree that allows for an accurate display of comparison between traditional and our approach.  Table \ref{tab:datasets} presents a brief summary of the data and statistics on the datasets used during our experiments.

\subsection{Doccano 1.0}
\label{sec:doccano_dataset}

After a thorough analysis of the available sources regarding subjective NLP tasks, we have noticed a certain lack of datasets focused on a variety of dimensions. Although part of the subjectivity area in NLP features one-coded labels of sarcasm or offensiveness, it may as well use a group of emotions, an example being Pluchik's wheel of emotions. However, none of them include a wide range of emotions, opinions, and feelings of the annotators. For this reason, we have launched a project named Doccano 1.0, where users annotated a diverse number of texts that corresponded to subjective problems. We have recruited around 40 individual people that were tasked to annotate 880 texts in the scope of 23 different subjective NLP tasks each. Each person annotated around 702 texts, and each text contains around 32 different annotations. It means that in total we acquired over 700k individual annotations.
Labels available for anotating were as follows: (1) \emph{positive}, (2) \emph{negative}, (3) \emph{joy}, (4) \emph{delight}, (5) \emph{inspiration}, (6) \emph{calm}, (7) \emph{surprise}, (8) \emph{compassion}, (9) \emph{fear}, (10) \emph{sadness}, (11) \emph{repulsion}, (12) \emph{anger}, (13) \emph{ironic}, (14) \emph{embarrassing}, (15) \emph{vulgar}, (16) \emph{political}, (17) \emph{interesting}, (18) \emph{understandable}, (19) \emph{incomprehensible}, (20) \emph{offensive to me}, (21) \emph{offensive to someone}, (22) \emph{funny to me} and (23) \emph{funny to someone}. Each of the 23 available labels had to be graded from 0 to 10, where 0 equals disagreement, and the latter a strong agreement.

\subsection{Unhealthy Conversations (UC)}
\label{sec:uc_dataset}

The Unhealthy Conversations (UC) dataset \cite{price2020six} was published in October 2020 and consists of 44k comments. Each piece of data can contain up to 250 characters sourced from Globe and Mail opinion articles that were sampled from the Simon Fraser University Opinion and Corpus dataset \cite{kolhatkar2020sfu}. The labels used to annotate these comments are as follows: (1) \emph{antagonize}, (2) \emph{condescending}, (3) \emph{dismissive}, (4) \emph{generalization}, (5) \emph{generalization unfair}, (6) \emph{healthy}, (7) \emph{hostile}, and (8) \emph{sarcastic}. There were at least three annotators per comment, and each text had to be described as at least one of the labels. Furthermore, to eliminate any possible bias, each comment was isolated from the context of the news articles and presented to the annotators as individual pieces of text.

\subsection{Measuring Hate Speech (MHS)}
\label{sec:mhs_dataset}

The Measuring Hate Speech (MHS) dataset \cite{kennedy2020constructing} is a collection of texts that were made available in 2020. It contains 39,565 comments from popular media platforms: Youtube, Twitter, and Reddit. The 7,912 Amazon Mechanical Turk workers specifically from the United States were involved in the annotation process. Since the data focus on offensiveness, it is possible to use labels that focus on different types of offensiveness, specifically: (1) \emph{disrespect}, (2) \emph{insult}, (3) \emph{humiliation}, (4) \emph{sentiment}, (5) \emph{attacking or defending nature of the post}, (6) \emph{dehumanization}, (7) \emph{inferiority of the status}, (8) \emph{hate speech}, (9) \emph{violence}, and (10) \emph{genocide}.
In our experiments, we treated each different type of offensiveness as a separate NLP task.

\begin{table*}
\centering
\caption{Dataset profiles after pre-processing. Each dataset contains a set number of labels, which are explained in full detail in Section \ref{sec:datasets}. The field \emph{Number of annotated labels} describes the number of annotations for all available labels in a specific dataset.}
\begin{tabular}{|l|c|c|c|}
\hline
\diagbox[width=15em]{Property}{Dataset} & Doccano 1.0 & Unhealthy Conversations & Measuring Hate Speech \\
\hline
Textual content profile & comments \& discussions & comments \& discussions & comments \\
\hline
\multirow{2}{*}{Number of tasks} & 23, i.e., delight, offensive to me, & 8, i.e. hostile, sarcastic,  & 10, i.e. violence,  \\
& funny to someone & unfair generalization & attack-defend, dehumanize \\
\hline
Label values & \{0, \ldots, 10\} & \{0, 1\} & \{0, \ldots, 4\} \\
\hline
Output / ML task & 23*regression & 8*binary classification & 10*regression \\
\hline
Number of texts & 1,000 & 44,355 & 39,565 \\
\hline
Number of annotations &  31,338  &  244,468 (227,975 valid) & 135,556 \\
\hline
Number of annotated labels & 720,774 & 1,823,800 & 1,355,560 \\
\hline
Number of annotators & 40 &  558 & 7,912 \\
\hline
Avg. annotations per text  & 31.34 & 4.66 & 3.43 \\
\hline
Avg. annotations per annotator & 783.45 & 387.71 & 17.13 \\
\hline
Language & Polish & English  & English \\
\hline
\end{tabular}
\label{tab:datasets}
\end{table*}

\section{Experimental Setup}
\label{sec:experimental_setup}

During our experiments, we used the 10-fold cross-validation (CV). In each iteration, we trained the model using 8 folds. One of the remaining folds was used as a validation set and another as a test set. Then, we calculated the mean and standard deviation of the evaluation metrics. In addition, to measure the significance of differences in the evaluation results between labels, we performed statistical tests. After verifying the test assumptions, we applied the \textit{t}-test for independent samples with Bonferroni correction. If the assumptions were not met, we used the Mann-Whitney \textit{U} test.

\subsection{Deep Neural Architectures}
\label{sec:deep_neural_architectures}

We fine-tuned the two variants of pre-trained transformer models, depending on the language of the data. In the case of the Doccano 1.0 dataset, we used the HerBERT model \cite{mroczkowski-etal-2021-herbert}. For the UC and MHS datasets, we leveraged XLM-RoBERTa \cite{conneau2019unsupervised}. The implementation of both models was obtained from the HuggingFace library \cite{wolf-etal-2020-transformers}. In both cases, we used the cross-entropy as a loss function with a learning rate set to $10^{-5}$. We leveraged the Adam optimizer with $\beta_1 = 0.9$, $\beta_2 = 0.999$ and the L2 regularization hiperparameter weight decay equals $10^{-3}$.

\subsection{Self-Supervised Model Evaluation}
\label{sec:self_supervised_model_evaluation}

\begin{figure*}[ht]
\centering
\includegraphics[width=\linewidth]{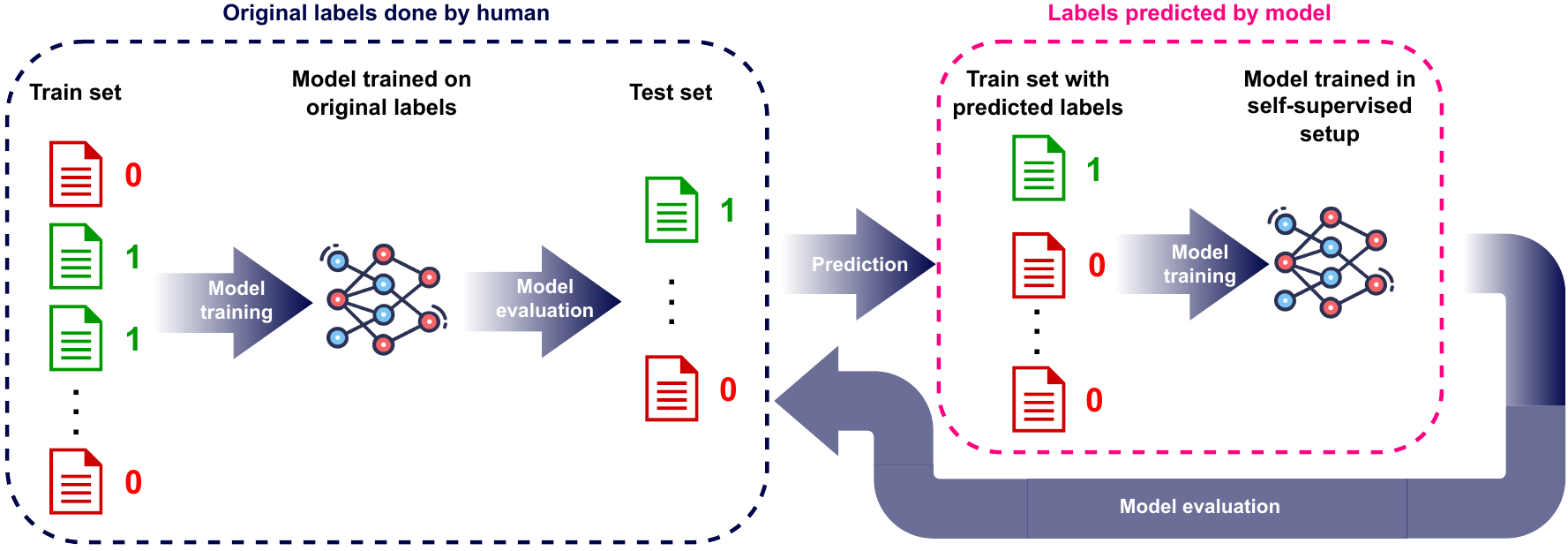}
\caption{The self-supervised evaluation scenario, where the model is trained on labels predicted by the instance trained on the original data, presented for a single label.} 
\label{fig:self_supervised_evaluation}
\end{figure*}

The first experiment was the self-supervised evaluation. We trained and evaluated the model on the original dataset. From the CV, we obtained the predicted labels for each text in the dataset. In the next step, we used them as target labels. In this way, we generated a second dataset in which text samples are exactly the same as in the original dataset, but target labels are predicted by the model. Subsequently, we trained the model on the predicted labels and evaluated it on the original set of labels. The self-supervised evaluation scenario is described in Fig.~\ref{fig:self_supervised_evaluation}.

\subsection{Incremental Knowledge Evaluation}
\label{sec:incremental_knowledge_evaluation}

Another experiment focused on increasing the train set size in an incremental way. We started with a training set consisting of only 1 fold for each iteration of the CV. Then, we increased the train set size to 2 folds, by adding the next fold to the one used in the previous experiment step. We kept increasing the train set size by 1 fold during the next steps. In the final iteration, we used 8 folds as a train set for each of the CV iterations.

\subsection{Threshold t Value Evaluation}
\label{sec:threshold_t_value_evaluation}

We also focused on the evaluation regarding various values of the $t$ threshold value used in the $VTL$ measure described in Sec.~\ref{sec:valuable_text_label}. To better understand the impact of the threshold $t$, we tested values in the range $[0.1, 0.25]$. In other words, we considered scenarios from the situation when we considered the label as valuable ($VTL(d,l,t)=1$) if at least 10\% of users annotated it with a non-zero value to the situation when the label needed to receive at least 25\% of non-zero annotations to be considered as valuable.

\subsection{Single-Task vs. Multi-Task}
\label{sec:single_task_vs_multi_task}

During the evaluation, we also analyzed the transfer of knowledge between tasks. Therefore, we measured the impact of inter-task knowledge by comparing the model performance in single-task and multi-task scenarios. 

\subsection{Diversity of People vs. Diversity of Texts}
\label{sec:diversity_people_vs_diversity_texts}

In this scenario, our goal was to investigate how many annotations from different people we should collect for one text to get the best quality of the model. In other words, we want to answer the question: \textit{Given a budget for (N) annotations, how many texts should we annotate?} To compare models trained on different combinations of annotation and text numbers, we used the following training dataset undersampling procedure: to create a training dataset with \( N\) annotations and \( M \) texts, we sort the texts by their number of annotations, descending. Then, we take top \( M \) texts, and sample their annotations in a round-robin way, until we get \( N \) annotations. The validation and test dataset splits remain fixed. We compare the model performance on the test dataset split.
We tested text numbers ranging from 100 to 500 in increments of 100, with annotation numbers ranging from 1,000 to 7,000 in increments of 1,000. 
An experiment was carried out for each combination on five test folds. We used a personalized UserID model with the HerBERT language model.

\section{Results}
\label{sec:results}

\begin{figure}
\centering
\includegraphics[width=\linewidth]{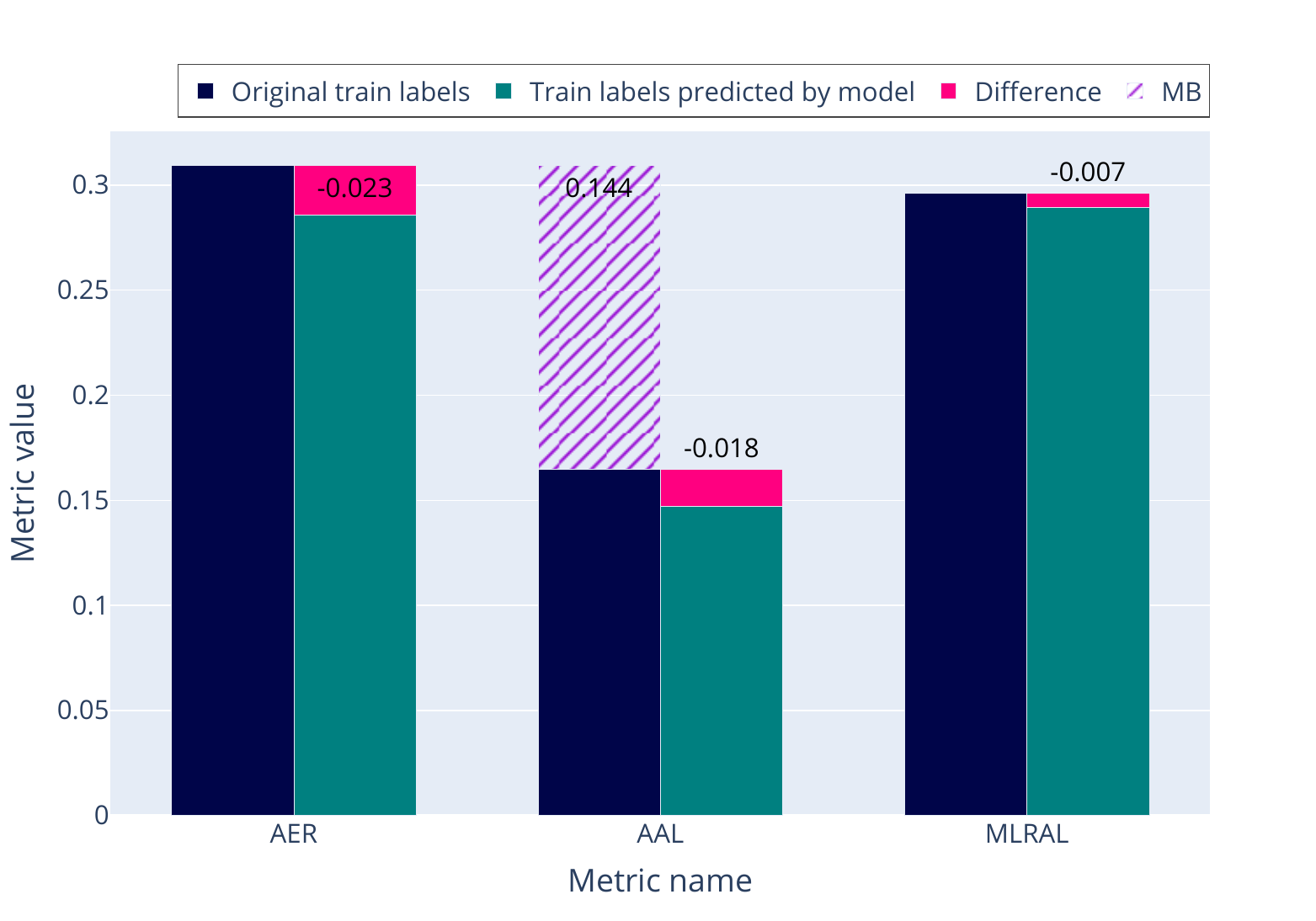}
\caption{The values of effort reduction AER, loss AAL, and MLRAL, as well as final model benefit MB for the self-supervised evaluation on the Doccano dataset.} 
\label{fig:self-supervised_doccano_our_metrics}
\end{figure}

\begin{figure}
\centering
\includegraphics[width=1.0\linewidth]{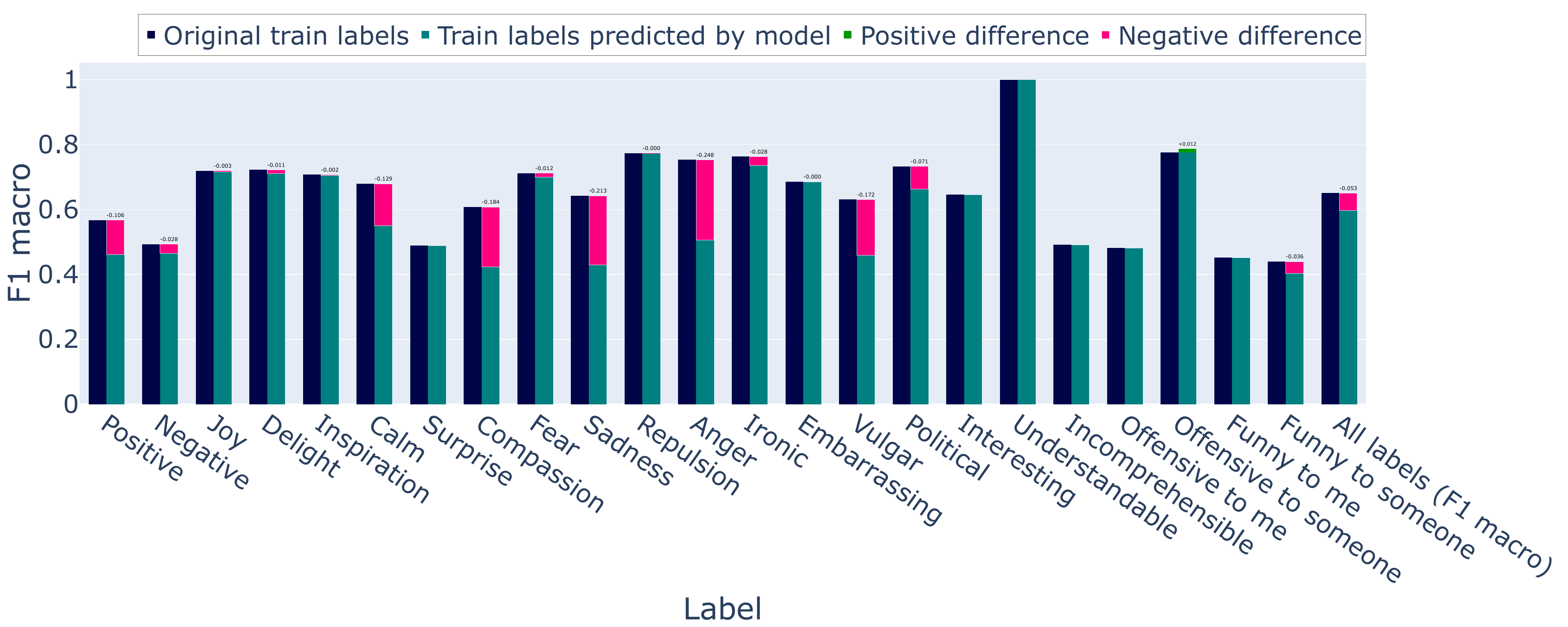}
\caption{Macro F1 values for the self-supervised evaluation on the Doccano dataset.} 
\label{fig:self-supervised_doccano_f1}
\end{figure}

In the case of self-supervised evaluation, the training on the labels predicted by the model resulted in a significant decrease in the model performance only for 7 out of 23 labels ($\sim$30\%) in comparison to the training on the original labels for the Doccano dataset described in Fig.~\ref{fig:self-supervised_doccano_our_metrics} and Fig.~\ref{fig:self-supervised_doccano_f1}. The \textit{Difference} marks the difference in performance between the model trained on the original labels and those predicted by the model. The shaded area marks the $MB$ metric that describes the reduction in annotation effort with respect to the loss of knowledge. 

\begin{figure}
\centering
\includegraphics[width=1.0\linewidth]{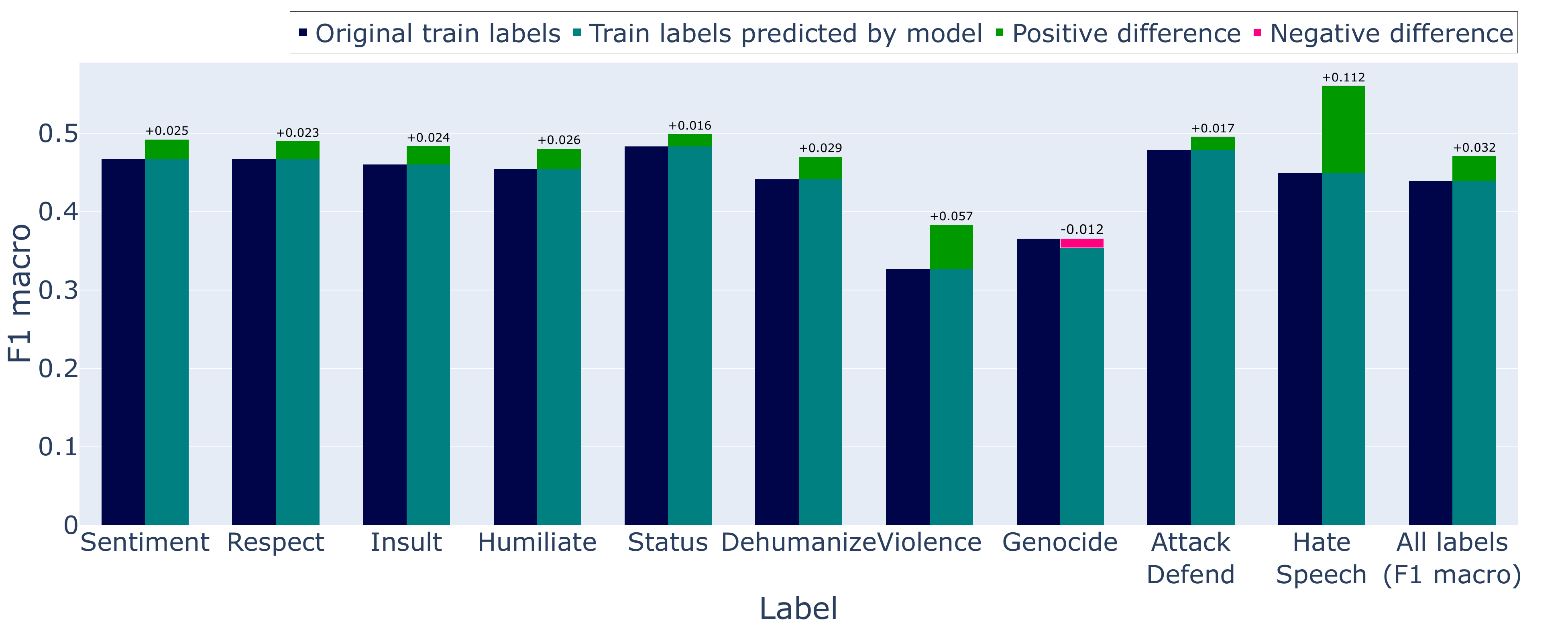}
\caption{Macro F1 values for the self-supervised evaluation on the MHS dataset.} 
\label{fig:self-supervised_mhs_f1}
\end{figure}

For the MHS dataset, the model trained in a self-supervised way achieved better results in 9 out of 10 labels (90\%), Fig.~\ref{fig:self-supervised_mhs_f1}. 

\begin{figure}
\centering
\includegraphics[width=1.0\linewidth]{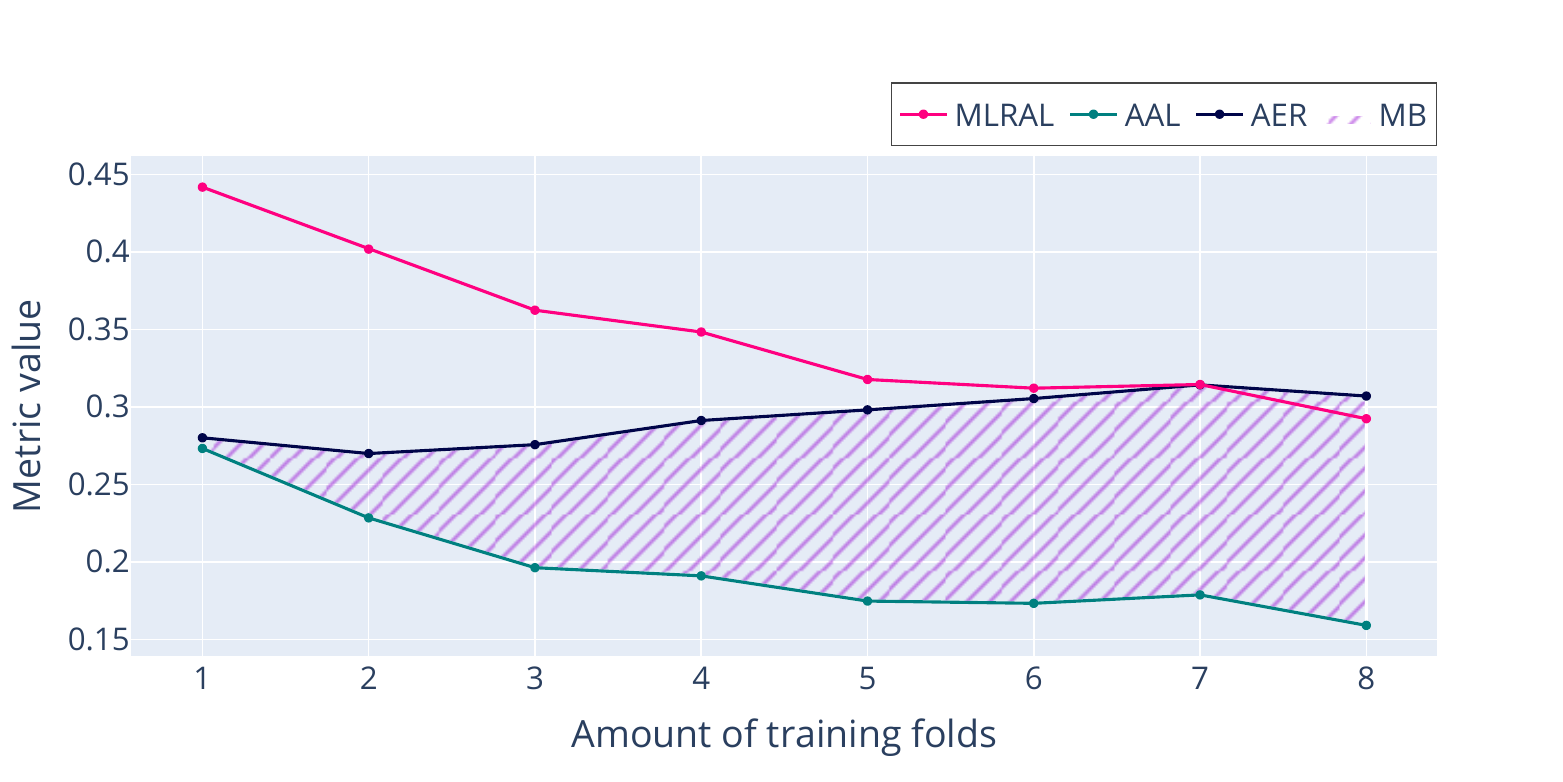}
\caption{The values of AER, AAL, and MLRAL metrics for the incremental knowledge evaluation on the Doccano dataset. The hatched area is Model Benefit (MB): the gain in effort reduction (AER) minus loss (AAL).} 
\label{fig:increment_doccano_our_metrics}
\end{figure}

For incremental knowledge evaluation, a positive $MB$ value was observed since training on 1 fold. The highest $MB$ value (0.15) was noted for training on 8 folds, Fig.~\ref{fig:increment_doccano_our_metrics}.

\begin{figure}
\centering
\includegraphics[width=1.0\linewidth]{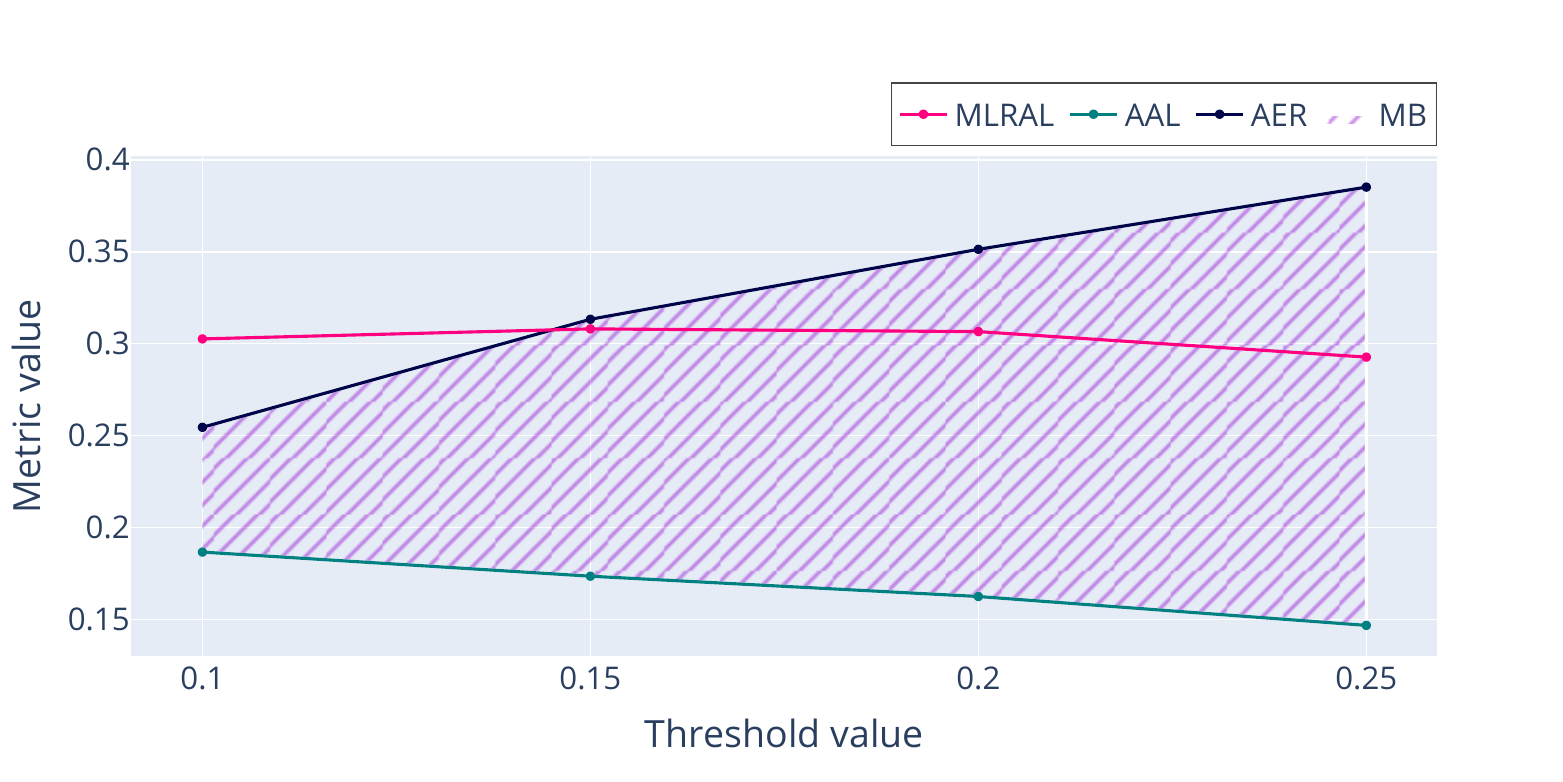}
\caption{The values of AER, AAL, and MLRAL metrics for the threshold evaluation on the Doccano dataset. The hatched area is Model Benefit (MB), i.e. the gain in effort reduction (AER) minus loss (AAL).} 
\label{fig:threshold_doccano_our_metrics}
\end{figure}

\begin{figure}
\centering
\includegraphics[width=1.0\linewidth]{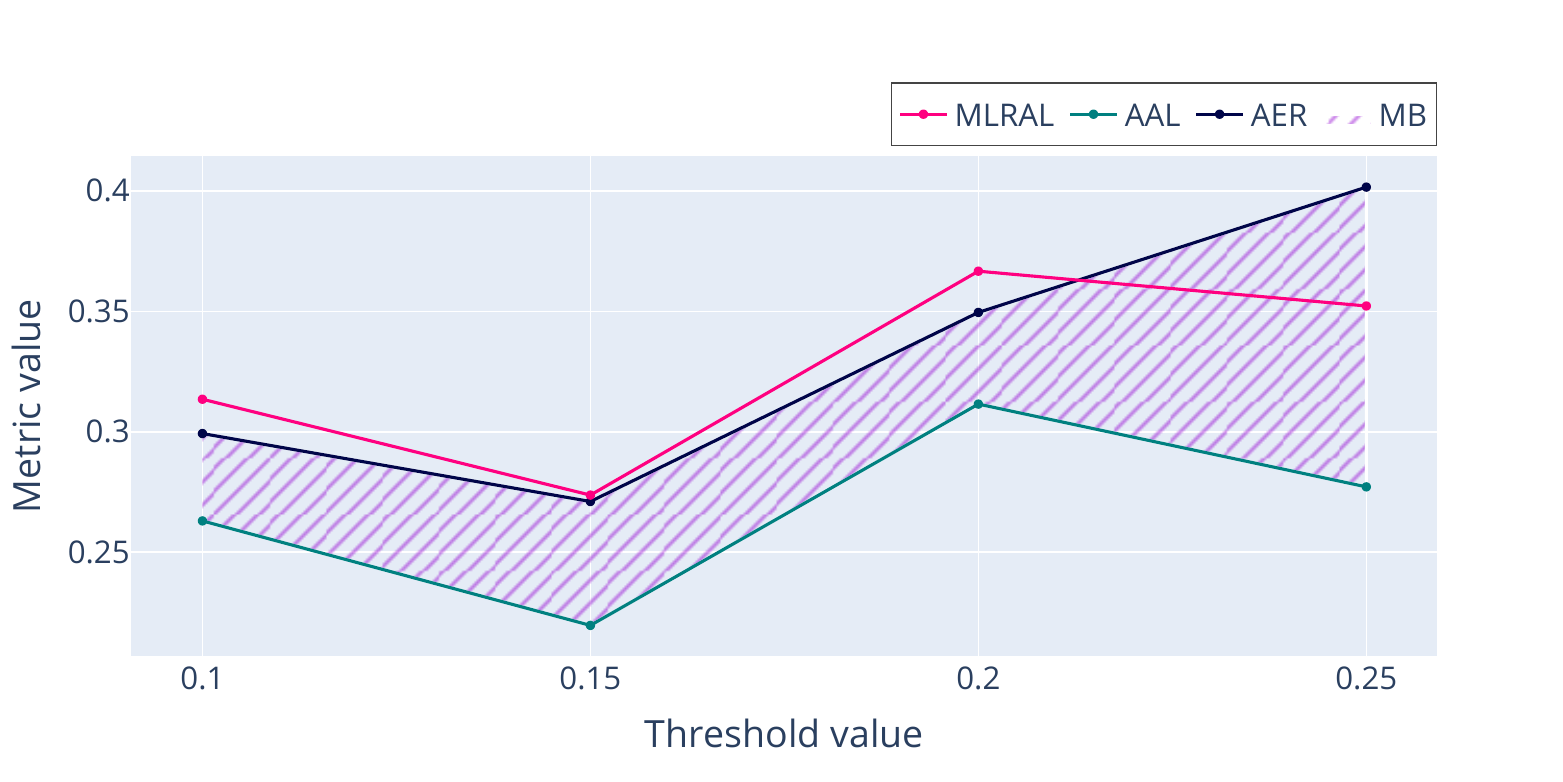}
\caption{The values of AER, AAL, and MLRAL for the threshold evaluation on the UC dataset. The hatched area is Model Benefit (MB): the gain in effort reduction (AER) minus loss (AAL).} 
\label{fig:threshold_uc_our_metrics}
\end{figure}

During the threshold $t$ value evaluation, the highest $AER$ and $MB$ values were observed for $t=0.25$ and were equal to 0.39 and 0.25, respectively, for the Doccano dataset, Fig.~\ref{fig:threshold_doccano_our_metrics}. For the UC dataset, the highest values for these metrics were equal to 0.41 and 0.14, Fig.~\ref{fig:threshold_uc_our_metrics}.

\begin{figure}
\centering
\includegraphics[width=1.0\linewidth]{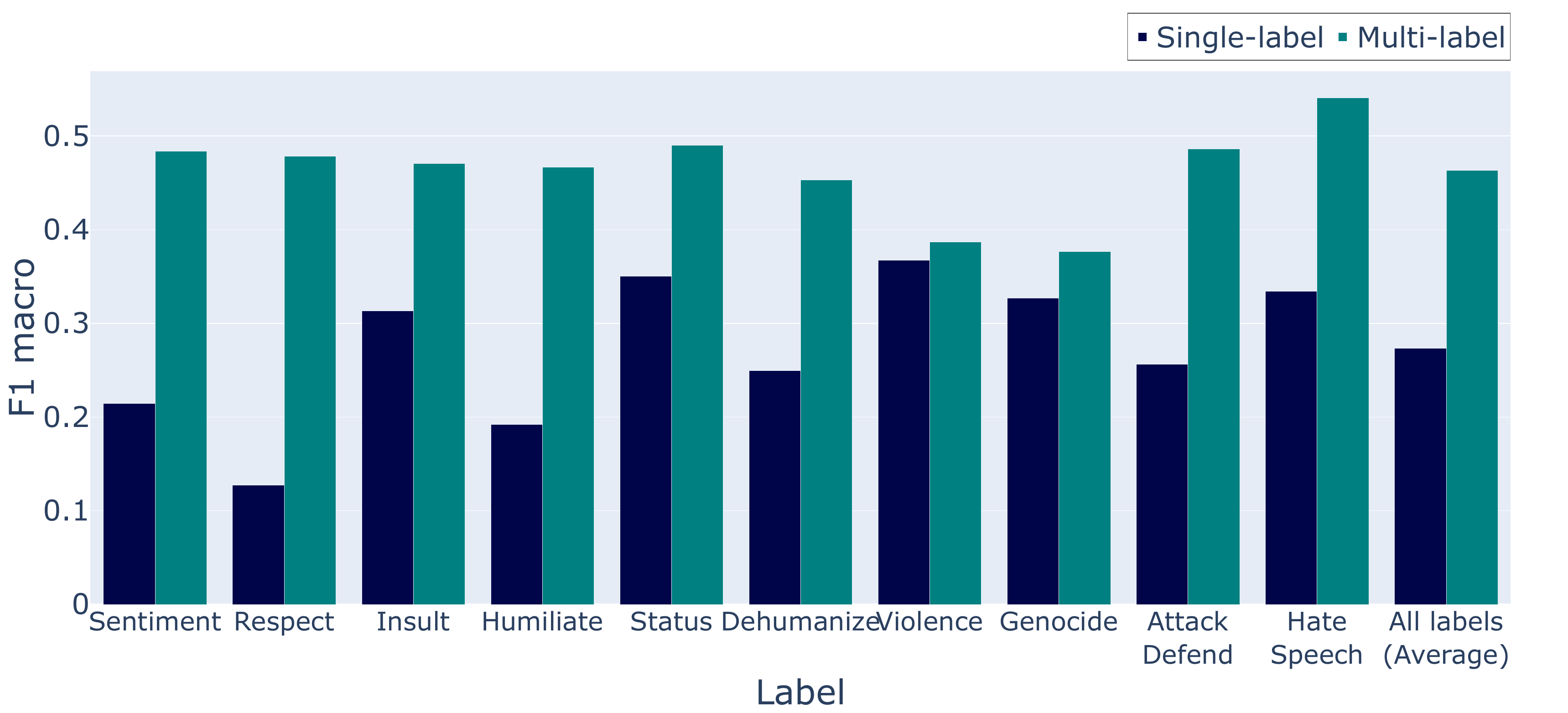}
\caption{Macro F1 values for the single-task vs. multi-task evaluation on the MHS dataset.} 
\label{fig:single_vs_multi_mhs_f1}
\end{figure}

For the single-task vs. multi-task evaluation, a significant increase in model performance was observed for every label in the MHS dataset, Fig.~\ref{fig:single_vs_multi_mhs_f1}. The highest increase caused by multi-task learning was 0.36 for \textit{Sentiment} label.

The results for the experiments on the dependence of the model quality on the number of texts in the training dataset are presented in Tab.~\ref{tab:doccano_performance_table}. Due to the limited number of annotations per text for the Doccano dataset, we were unable to evaluate some combinations of the number of texts and annotations (e.g. 100 texts and 7,000 annotations). A slightly positive effect of more texts on the quality of the model is visible. This may be due to the fact that linguistic knowledge is more important than personalization in the case of small datasets.

\begin{table}
\centering
\caption{Model performance for different text/annotations number for Doccano 1.0 dataset. We used averaged \(R^2\) over all dataset tasks as a metric.}
\begin{tabular}{|c|c|c|c|c|c|}
\hline
\diagbox{Annotations \\ number}{Texts \\ number} &    100 &    200 &    300 &    400 &    500 \\
\hline
1,000               &   08.34 &   09.59 &  11.16 &  12.77 &  10.73 \\
\hline
2,000               &  19.24 &  17.46 &  21.34 &  22.86 &  22.75 \\
\hline
3,000               &    --    &  23.73 &  24.96 &  27.31 &  26.10 \\
\hline
4,000               &    --    &  25.00 &  27.75 &  28.65 &  28.05 \\
\hline
5,000               &   --     &     --   &  28.04 &  29.07 &  29.59 \\
\hline
6,000               &     --   &     --   &  29.05 &  30.13 &  29.72 \\
\hline
7,000               &   --     &    --    &     --   &  30.22 &  29.96 \\
\hline
\end{tabular}
\label{tab:doccano_performance_table}
\end{table}

\section{Discussion}
\label{sec:discussion}

Behind every performance of NLP methods, there is the data and the significant impact of its quality. Even the slightest noise may hinder the model performance, and thus it is important to focus on the characteristics of our data. The number of positive and negative values of a single label in our experiments did not have balanced proportions, since the negative values were in the dominant part of each class. As a result, there is a certain bias towards negative values, which may aggravate the model predictions. This is a factor that adversely affects our methods, as the loss of a small number of minority class annotations could significantly reduce the quality of the model performance. Therefore, the high MB values and resulting loss-adjusted benefits of our methods confirm their versatility and robustness to unfavorable annotation distributions. 
On the other hand, we think that a re-evaluation of the scores with a technique that includes other features and values of each label (i.e., confusion matrix) would be applicable to further improve the stability of our methods.

The very similar trend observed for MLRAL and AAL for the Doccano dataset indicates that the loss of valuable (non-zero) annotations is independent of the diversity of annotation distributions for individual tasks. This reveals the prospect of using our methods effectively on datasets annotated with multiple tasks, regardless of their distributions.

Just as the value of the loss-adjusted MB measure is relevant for each collection used, we are fully aware that the objective of annotation may vary depending on the characteristics of the phenomenon being annotated. This means that in special cases, the loss of valuable annotation may prevent the model from extracting the correct signal from the data. That is when MLRAL minimization may prove more important than MB maximization. On the other hand, for a phenomenon that occurs quite frequently, maximization of AER may prove more efficient.

The low performance differences in the self-supervised scenario on the Doccano dataset indicate that the model stores much of the knowledge necessary to predict on unknown texts in a way that does not significantly reduce the quality of the labels it provides. This makes it possible to incrementally enlarge the dataset to, for example, support more domains or adapt the model to another language.

The improvement in performance in the MHS dataset in the self-supervised scenario as shown in Fig.~\ref{fig:self-supervised_mhs_f1} is due to the fact that training the model on the predicted labels can be treated as a regularization method.

The use of inter-task knowledge has significantly improved the model performance on the MHS set in the single-task vs. multi-task scenario. By simultaneously predicting VTL values for each task, the model learns inter-task relationships, which allows it to discover the hidden semantics of the tasks it learns.

\section{Conclusions and Future Work}
\label{sec:future_work}
In this work, we proposed a new model-based approach to data acquisition for subjective multi-task NLP problems. We develop novel metrics to calculate the benefit and loss of our method. We also performed a complex evaluation to verify the efficiency of the proposed solution from multiple perspectives. One of the most important results of our experiments is that the self-supervised approach can be used as a regularization technique for subjective multi-task problems. With this setup, the model is able to remove outliers and subsequently improve the overall performance quality. The use of our approach can lead to a reduction in the overall annotation effort by 40\%. Assuming the pricing in annotation services such as Amazon Mechanical Turk\footnote{\url{https://aws.amazon.com/sagemaker/data-labeling/pricing/}}, the cost to annotate one label is \$0.012. On that basis, we can calculate that the cost of annotating the data set with a size comparable to the UC dataset would be \$219k, but the 40\% reduction caused by our method would allow saving up to \$87.6k. 

In future work, we will focus on further improvements of our model-based approach by using different model architectures, developing new measures tailored for specific business cases, and applying our method to more datasets regarding subjective multi-task problems.
The code for all methods and experiments is publicly available \footnote{\url{https://github.com/CLARIN-PL/model-based-data-acquisition/releases/tag/2023-icdm-sentire}} under the MIT license.

\section*{Acknowledgements}
This work was financed by 
(1) the National Science Centre, Poland, project no. 2021/41/B/ST6/04471;  
(2) Contribution to the European Research Infrastructure 'CLARIN ERIC - European Research Infrastructure Consortium: Common Language Resources and Technology Infrastructure', 2022-23 (CLARIN Q);
(3) the Polish Ministry of Education and Science, CLARIN-PL; 
(4) the European Regional Development Fund as a part of the 2014-2020 Smart Growth Operational Programme, projects no. POIR.04.02.00-00C002/19, POIR.01.01.01-00-0923/20, POIR.01.01.01-00-0615/21, and POIR.01.01.01-00-0288/22;  
(5) the statutory funds of the Department of Artificial Intelligence, Wroclaw University of Science and Technology;
(6) the Polish Ministry of Education and Science within the programme “International Projects Co-Funded”;
(7) the European Union under the Horizon Europe, grant no. 101086321 (OMINO). However, the views and opinions expressed are those of the author(s) only and do not necessarily reflect those of the European Union or the European Research Executive Agency. Neither the European Union nor European Research Executive Agency can be held responsible for them.

\bibliographystyle{IEEEtran}
\bibliography{ecai, main}

\end{document}